\documentclass{article}

\usepackage{microtype}
\usepackage{graphicx}
\usepackage{subfigure}
\usepackage{booktabs} 

\usepackage{tabularray}
\UseTblrLibrary{booktabs}
\usepackage{pifont}

\usepackage{hyperref}

\usepackage[accepted]{icml2025}

\usepackage{amsmath}
\usepackage{amssymb}
\usepackage{mathtools}
\usepackage{amsthm}

\usepackage[capitalize,noabbrev]{cleveref}

\theoremstyle{plain}
\newtheorem{theorem}{Theorem}[section]
\newtheorem{proposition}[theorem]{Proposition}

\theoremstyle{definition}

\theoremstyle{remark}

\newcommand\etc{\textit{etc.}}

\icmltitlerunning{BlueGlass}

\begin{document}

\twocolumn[
\icmltitle{BlueGlass: A Framework for Composite AI Safety }



\begin{icmlauthorlist}
\icmlauthor{{Harshal Nandigramwar}}{intel,unistu}
\icmlauthor{Syed Qutub}{intel}
\icmlauthor{Kay-Ulrich Scholl}{intel}
\end{icmlauthorlist}

\icmlaffiliation{intel}{Intel Labs, Munich, Germany}
\icmlaffiliation{unistu}{University of Stuttgart, Stuttgart, Germany}

\icmlcorrespondingauthor{Syed Qutub}{syed.qutub@intel.com}

\icmlkeywords{Vision-Language Models, Explainable AI, Interpretability, Object Detection, ICML}

\vskip 0.3in
]



\printAffiliationsAndNotice{}  

\begin{abstract}

As AI systems become increasingly capable and ubiquitous, ensuring the safety of these systems is critical. However, existing safety tools often target different aspects of model safety and cannot provide full assurance in isolation, highlighting a need for integrated and composite methodologies. This paper introduces \textsc{BlueGlass}, a framework designed to facilitate composite AI safety workflows by providing a unified infrastructure enabling the integration and composition of diverse safety tools that operate across model internals and outputs. Furthermore, to demonstrate the utility of this framework, we present three safety-oriented analyses on vision-language models for the task of object detection: (1) distributional evaluation, revealing performance trade-offs and potential failure modes across distributions; (2) probe-based analysis of layer dynamics highlighting shared hierarchical learning via phase transition; and (3) sparse autoencoders identifying interpretable concepts. More broadly, this work contributes foundational infrastructure and findings for building more robust and reliable AI systems.

\end{abstract}

\section{Introduction}
\label{sec:intro}

\begin{figure*}[ht]
\begin{center}
\includegraphics[width=\textwidth]{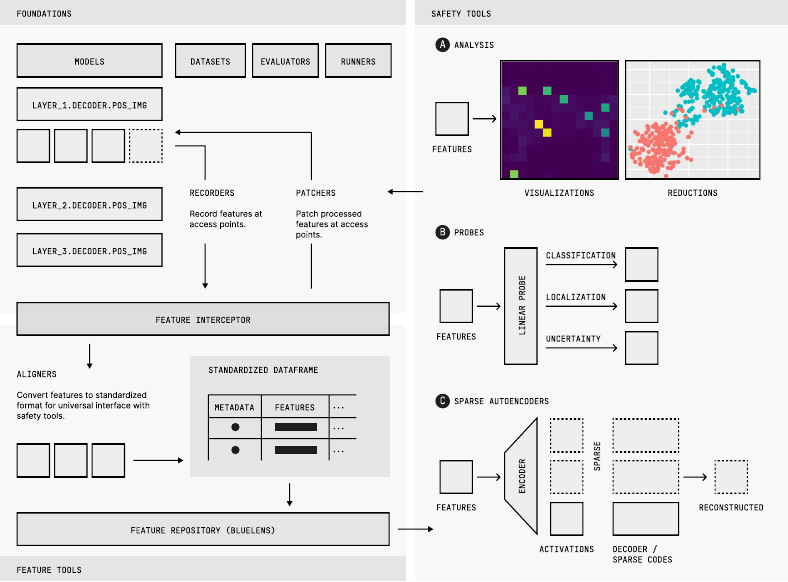}
\caption{System overview of {\textsc{BlueGlass}}, a framework for composite AI safety.}
\label{fig:overview}
\end{center}
\end{figure*}

Ensuring the safety and reliability of complex AI systems presents significant challenges, as no single tool or method can provide the required level of assurance~\cite{composite_neel, composite_adversarial, composite_saliency}. Existing safety and interpretability approaches often target complimentary aspects of safety in isolation and have inherent limitations~\cite{negative_saes, composite_adversarial, composite_saliency}, making them insufficient as such. Addressing comprehensive AI safety necessitates an ensemble of diverse tools that can be integrated and composed to cover each other's weaknesses – a notion we refer to as composite AI safety.

In this paper, we introduce \textsc{BlueGlass}, an open source framework designed to facilitate the composite AI safety methodology by enabling the integration and composition of diverse safety tools operating across model internals and outputs within a unified infrastructure. To demonstrate the various aspects of this framework, we conduct analyses in form of case studies to analyze vision-language models (VLMs)~\cite{vlm_survey}, which are rapidly rising in prominence, and in particular for the task of object detection~\cite{odsurvey} due to it's potential in general-purpose robotics~\cite{gemini_robotics, robo_survey} and autonomous driving~\cite{auto_survey} where reliable object detection is a crucial requirement.

We present three safety-oriented analyses on VLMs for object detection, yielding several key contributions. Firstly, through empirical safety analysis via distributional evaluation, we uncover key performance trade-offs and potential failure modes of state-of-the-art VLMs across diverse operational scenarios~(\cref{sec:evals}). Secondly, we propose a variant of linear probes~\cite{linearprobes}, called \textit{approximation probes}, to analyze the layer dynamics of VLMs and vision-only detectors, revealing a universal phase transition~\cite{pt_dlst, pt_induction, pt_double_descent} phenomenon indicative of shared hierarchical feature learning principles~(\cref{sec:features}). Finally, using sparse autoencoders~\cite{sae_ksparse, sae_features}, we perform concept decomposition and discovery, identifying interpretable concepts and highlighting spurious correlations learned by VLMs through dataset attribution~(\cref{sec:concept}).

In essence, this work contributes open source infrastructure for composite AI safety research and provides empirical and mechanistic insights into VLM capabilities, limitations, and internal representations, informing safety-aware deployment and future research.

\section{System Design and Principles}
\label{sec:principles}

Modern AI safety research draws on a wide range of techniques, from mechanistic analysis/intervention of model internals~\cite{mechinterp, sae_refusal, sae_steering} to robustness assessments~\cite{evals_robust} and data attributions~\cite{evals_attrib}. These methods operate at different layers of abstraction and require varying forms of access to a model’s structure, behavior, and computations. At the same time, the underlying models continue to evolve in architectural complexity, heterogeneity, and deployment scale~\cite{survey_diffusion, survey_transformers, survey_state_space}.

To support the development and integration of these safety tools, we introduce a unified infrastructure designed for \textit{generality}, \textit{composability}, \textit{resourcefulness}, and \textit{usability} as further described in~\cref{appx:principles}. The framework comprises of three core abstractions that enables building and composing of safety tools that operate over model internals, outputs, and evaluation metrics, all within a common execution and data management framework as shown in~\cref{fig:overview}. 

\subsection{Foundations} 
\label{sec:foundations}

This layer provides the essential building blocks that the framework operates upon and interacts with. It includes modules for interfacing with diverse \textit{models}, managing various \textit{datasets}, defining and executing \textit{evaluators} for performance and safety evaluations, and orchestrating experimental runs via \textit{runners}. These components provide an abstraction over various sources such as HuggingFace~\cite{huggingface}, detectron2~\cite{detectron}, mmdetection~\cite{mmdetection}, and custom implementations, so that they can be readily integrated with other components through a unified interface. 

\subsection{Feature Tools}
\label{sec:features}

Building on the advances in mechanistic interpretability~\cite{mechinterp}, recent AI safety research increasingly focuses towards intrinsic methods that inspect, monitor and manipulate internal representations of models~\cite{actpatch, sae_steering, sae_features}. However, existing infrastructure to manage these artifacts is often tightly coupled to specific architectures~\cite{transformerlens, saelens}, inaccessible~\cite{garcon}, restrictive for complex code bases~\cite{nnsight} or poorly integrated across workflows. To enable a principled and unified interface across diverse tasks, model architectures and safety methods, we propose a system for model internals management within the \textsc{BlueGlass} framework.

\paragraph{Interceptor.} This is the central module that enables access to model internals. This component wraps a target model, providing a standardized mechanism to define access points within the model's architecture where features can be captured or modified. The \textit{interceptor} supports both \textit{manual mode}, where users explicitly insert calls in model code, and a \textit{hooked mode}, which automates access for standard architectures based on a mapping of abstract access points to model layers.

\paragraph{Recorder.} Working in conjunction with the \textit{interceptor}, \textit{recorders} are responsible for capturing feature artifacts at the designated access points during a model's execution. They provide the mechanism to extract intermediate representations for subsequent analysis workflows.

\paragraph{Patcher.} In complement to the \textit{recorder}, \textit{patchers} allow for the modification of feature artifacts at access points. This capability is essential for performing interventions such as activation patching~\cite{actpatch}, steering~\cite{sae_steering}, or other feature manipulation experiments, enabling counterfactual analysis of model behavior.

\paragraph{Aligner.} Once the raw features are captured, \textit{aligners} process them to ensure consistency. Given the inherent heterogeneity in feature artifacts with shapes and formats across different models and layers, \textit{aligners} transform these features into a standardized format conforming to a defined schema. This step is enables downstream compatibility with other components in the system, ensuring they can operate on features from diverse sources uniformly.

\paragraph{Storage.} To reduce the runtime cost of intrinsic analyses, the processed features in the standard schema are stored to disk and are loaded in an efficient manner as per the requirement. The storage utilizes Apache Arrow~\cite{apachearrow} tables and Parquet~\cite{apacheparquet} format, which are optimized for efficient columnar storage and high-performance data loading. A \textit{FeatureDataset} wrapper provides a convenient interface for accessing the stored features, streamlining data loading for analysis tasks such as training probes or sparse autoencoders. Furthermore, the system supports streaming data access through integration with the HuggingFace~\cite{huggingface} datasets platform, enhancing flexibility for large-scale data processing.

The \textit{feature tools} layer provides a unified, flexible, and efficient system for managing the internal representations of diverse models. By abstracting the complexities of feature access, standardization, and storage, these tools establish a critical foundation for conducting various white-box AI safety analyses.

\subsection{Safety Tools}
 Building upon the core foundations~(\cref{sec:foundations}) and the robust model internals management system~(\cref{sec:features}), the framework empowers researchers to compose and deploy a diverse array of AI safety tools. These tools seamlessly interact with target models and datasets via the \textit{foundations} and utilize the standardized access to internal representations provided by the \textit{feature tools} to enable composite AI safety workflows. The capabilities and practical applicability of the framework in supporting safety workflows are demonstrated in the following sections through detailed case studies focusing on safety-oriented evaluation methodologies~(\cref{sec:evals}), probing of representations~(\cref{sec:probes}), and concept analysis using sparse autoencoders~(\cref{sec:concept}) for vision-language models on the task of object detection. 
\section{Empirical Safety Analysis via Distributional Evaluations}
\label{sec:evals}

\begin{table*}[ht]
\scriptsize
\resizebox{\textwidth}{!}{%
\noindent
\begin{tblr}{%
  hspan=minimal,
  colsep=2.8pt,
  cell{1}{1} = {r=2}{},
  cell{1}{2} = {c=4}{},
  cell{1}{6} = {c=2}{},
  cell{1}{8} = {c=2}{},
  cell{1}{10} = {c=2}{},
  cell{1}{12} = {c=2}{},
  cell{1}{14} = {c=2}{},
  cell{1}{16} = {c=2}{},
  cells = {c, m},
  column{1} = {l, m},
}
\toprule
\textbf{Method}  & 
\textbf{\tiny Attributes} & & & &
\textbf{\tiny FunnyBirds} & &
\textbf{\tiny ECPersons} & & 
\textbf{\tiny Valerie22} & & 
\textbf{\tiny BDD100k} & &
\textbf{\tiny COCO} & & 
\textbf{\tiny LVIS} &  \\
\cmidrule[lr]{2-5} 
\cmidrule[lr]{6-7} 
\cmidrule[lr]{8-9} 
\cmidrule[lr]{10-11} 
\cmidrule[lr]{12-13}
\cmidrule[lr]{14-15}
\cmidrule[lr]{16-17}
& Type & Box & Size & FPS & 
AP & AR & 
AP & AR & 
AP & AR & 
AP & AR & 
AP & AR & 
AP & AR \\
\midrule
{YOLO v8 \\ \tiny\cite{yolo}} & D & \checkmark & 0.068 & 71.5 &
85.2 &  95.4  & 1.1 & 31.1 & 1.1 & 38.5 & 8.8 & 19.4 & 24.9 & 42.6 & 7.1 & 14.1 \\
\midrule
{Grounding DINO \\ \tiny\cite{gdino}} & C & $\checkmark$ & 0.172 & 8.3 &
87.3 &  91.2  & \underline{22.1} & \underline{46.5} & \underline{15.2} & \underline{50.8} & \underline{23.8} & \underline{\textbf{59.4}} & \underline{48.5} & \underline{77.2} & 14.2 & 53.2 \\
{GenerateU \\ \tiny\cite{genu}} & G & $\checkmark$ & 0.896 & 1.5 &  
65.1 &  92.9  & 2.4 & 34.6 & 2.1 & 42.6 & 13.1 & 37.7 & 32.1 & 66.1 & \underline{\textbf{25.5}} & \underline{\textbf{40.7}} \\
{Florence 2 Large \\ \tiny\cite{florence}} & G & $\times$ & 0.822 & 2.9 & 
\underline{87.9} & \underline{93.0}  & 1.6 & 30.7 & 1.3 & 43.5 & 11.7 & 25.5 & 40.1 & 55.2 &  2.3 & 0.3 \\
\midrule
{Gemini 2.0 Flash \\ \tiny\cite{gemini}} & G & $\times$ & \dag & \dag & 
32.2 & 50.0  & 1.3 & 21.3 & 0.1 & 15.7 & 0.9 & 3.4 & 19.9 & 32.8 & 4.9 & 7.2 \\
\midrule
{DINO (SFT) \\ \tiny\cite{detr}} & D & $\checkmark$ & 0.218 & 4.8 &
\textbf{99.6} & \textbf{99.9} & \textbf{66.4} & \textbf{76.0} & \textbf{37.4} & \textbf{70.2} & \textbf{35.9} & 55.6 & \textbf{58.3} & \textbf{78.6} & 20.8 & 38.7 \\
\bottomrule
\end{tblr}}
\caption{Evaluation results for vision-language models on object detection. Attribute \textit{type} represents the VLM architecture type among discriminative (D), contrastive (C) and generative (G)~(see~\cref{appx:vlm_arch}); \textit{FPS} stands for frames per second measuring the runtime performance; \textit{box}, highlights inclusion of proposal information in VLM architecture; \textit{size} is the model parameter count in billions.}
\label{tab:eval}
\end{table*}

Evaluations serve as the first tool of choice in the AI safety toolkit~\cite{evals_behave}, providing empirical evidence necessary for identifying model behaviors, understanding capability boundaries, and assessing performance under varying conditions. 

Vision-language models (VLMs) have emerged as a promising solution to general-purpose robotics~\cite{gemini_robotics, robo_survey} and autonomous driving~\cite{auto_survey} where object detection is a crucial component. Hence, it is crucial to investigate its capabilities and potential limitations to ensure safe deployment. This section presents an evaluation of sate-of-the-art VLMs on the task of object detection, conducted as a safety analysis, investigating the behavior of VLMs across diverse conditions to identify failure modes relevant to its deployment.

\subsection{Experiment Setup}

\paragraph{Models.} Our evaluation encompasses six distinct architectures, selected to represent a broad perspective on the current state of object detection performance and capabilities of vision-language models.

To provide reference points for VLM performance within the object detection task, we utilize the zero-shot evaluation of pretrained YOLOv8x~\cite{yolo} on OpenImages v7~\cite{openimages} as a baseline. In complement, to establish an upper bound representative of strong performance achievable when models are fine-tuned on the target datasets, we include results from a DINO-DETR~\cite{detr} (Swin-L backbone) fine-tuned on respective datasets with AdamW ($\beta_1=0.9$, $\beta_2=0.999$, $lr=1\times10^{-4}$), batch size of 16 and default configuration.

For vision-language models, which are the primary subject of this evaluation, we group them into three 
architectural classes, as described in~\cref{appx:vlm_arch}, and evaluate the representative, best-performing models from each category. These include Grounding DINO~\cite{gdino}, GenerateU~\cite{genu}, and Florence 2~\cite{florence} with their characteristics described in~\cref{tab:eval}. Furthermore, we include results for Gemini 2.0 Flash~\cite{gemini}, representing a prominent closed-source VLM. Notably, although we explored other VLMs, such as LLaVA-NeXT~\cite{llava}, GPT 4o-mini~\cite{gptmini}, and PaliGemma-2~\cite{paligemma}, \etc, their outputs were unparsable or incorrect, leading to their exclusion.

\paragraph{Datasets.} Distributional evaluations are solely characterized by the choice of datasets. For this evaluation we assess the performance of VLMs on datasets that represent diverse operational scenarios and data characteristics. This includes varying sizes of label set, varying deployment scenarios (autonomous driving with Eurocity Persons~\cite{ecpersons} and BDD100k~\cite{bdd}, common objects with COCO~\cite{coco} and LVIS~\cite{lvis}, \etc), open and closed set settings (with LVIS~\cite{lvis} and others), contrasting data distributions (synthetic with ECPersons~\cite{ecpersons} and real with VALERIE22~\cite{valerie}) and out-of-distribution performance, with creation of a synthetic object detection dataset, based on FunnyBirds~\cite{funnybirds}.

\paragraph{Evaluation.} VLMs present a unique challenge in evaluation compared to traditional object detectors. This is primarily because of their open-ended textual outputs instead of predictions from a fixed vocabulary. To address this issue and ensure standardized, fair comparison across model types, we design a pipeline, as shown in~\cref{fig:eval_pipeline}, that maps the open-ended predictions of VLMs to the dataset-specific classes and incorporates scores to generate standardized predictions. A detailed description of the pipeline's components, methodology and ablations validating it's design are presented in~\cref{appx:eval_ablation}. For metrics, we report average precision (AP) and average recall (AR), which are a standard for object detection. For datasets that do not recommend an evaluation scheme or produce non-standard metrics, we use the COCO evaluation scheme~\cite{coco}.

\subsection{Results}

The evaluation results presented in~\cref{tab:eval} demonstrate the trade-offs between VLMs and vision-only closed-set detection baselines, along with their current limitations.

\paragraph{Zero-shot capabilities.} YOLOv8 establishes a strong baseline for zero-shot performance. Despite being a closed-set model repurposed as a zero-shot baseline, YOLOv8 (7.1) outperforms Florence (2.3) and Gemini (4.9) on open-vocabulary detection. This challenges the VLMs' readiness for general-purpose detection. This discrepancy stems from the model's lack of localization components that induce geometric priors. Grounding DINO (14.2) and GenerateU (25.5) perform better compared to other VLMs, but only GenerateU performs better than the fine-tuned DINO (20.8) baseline. In general, all VLMs demonstrate decent zero-shot performance across datasets.

\paragraph{Comparison with fine-tuned baselines.} Fine-tuned DINO outperforms all other models in all cases except in open-vocabulary detection. This establishes the relevance of supervised models in domain-specific tasks. Notably, on EuroCity Persons and Valerie22, DINO achieves 2-3 times better accuracy, demonstrating the gap between VLM capabilities and the supervised counterpart for dense and fine-grained object detection. However, the computational cost of per-dataset tuning and the availability of supervised datasets limits broader applicability, a weakness VLMs aim to address through their generalization capabilities. 

\paragraph{Open-ended detection.} VLMs excel in tasks requiring open-ended semantic reasoning. In particular, the combination of a detection network and a language model, as in GenerateU, balances the semantic reasoning capabilities and geometric priors to achieve best performance on open-vocabulary benchmark. Notably, Grounding DINO, a contrastive VLM, still struggles on this benchmark because of limited expressivity of contrastive objective.

\paragraph{Domain Generalization.} The consistent AR scores across various VLMs on the FunnyBirds dataset suggest comparable performance and good generalization capabilities. However, GenerateU exhibits a drop in AP, likely due to the lack of grounding in its output space. This observation underscores the necessity of incorporating grounding mechanisms into VLMs to achieve better controllability over their outputs.

\paragraph{Architectural trade-offs in VLMs.} Object detection with VLMs presents a trade-off between semantic capabilities, spatial capabilities, task diversity and computational cost. Contrastive VLMs (Grounding DINO), that utilize contrastive heads for classification, prioritize resource efficiency and perform well on small-to-medium label sets but sacrifice semantic capabilities and task diversity, which is evident in their open-vocabulary detection performance. Conversely, generative VLMs, either with (GenerateU) or without (Florence, Gemini) explicit detection components, trade-off computational cost with task diversity and spatial capabilities respectively. The need for geometric priors, through mechanisms like detection networks or improved architectures, is apparent for precise object localization, a capability crucial for enhancing visual understanding in VLMs.
\section{Approximation Probes for Intrinsic Analysis of Layer Dynamics}
\label{sec:probes}

\begin{figure}[ht]
\begin{center}
\centerline{\includegraphics{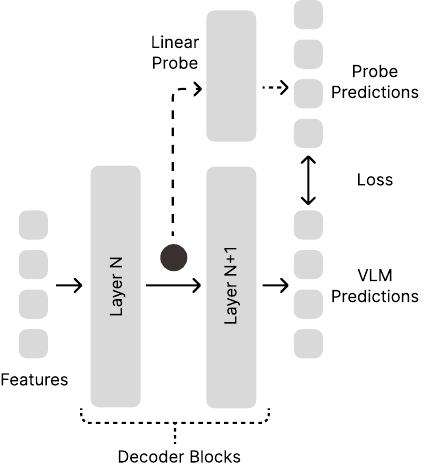}}
\caption{Linear probes setup on decoder layers.}
\label{fig:probes}
\end{center}
\end{figure}

\begin{figure*}[ht]
\begin{center}
\centerline{\includegraphics{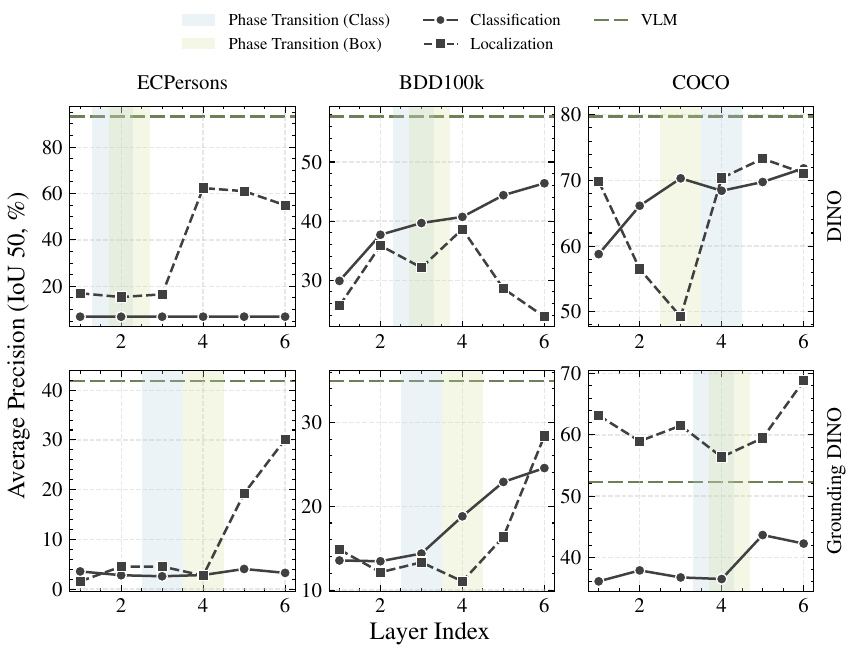}}
\caption{Phase transition in decoder layers of closed-set detector and vision-language model. }
\label{fig:phase_transition}
\end{center}
\end{figure*}

Representation probing~\cite{linearprobes} is a class of AI safety methods that extend the target model with light-weight linear layers to quantitatively measure auxiliary information, such as information content, uncertainties, \etc, at intermediate model positions. Recent works have shown their applicability for a wide range of safety related tasks such as mechanistic analysis~\cite{probe_physics}, improving robustness~\cite{probe_robust}, model debugging~\cite{probe_debug} \etc, and hence we include support for commonly used probes within the \textsc{BlueGlass} framework. Further, we propose a novel variant of linear probes~\cite{linearprobes} called \textit{approximation probe}, and demonstrate it's applicability to study layer dynamics, in this section.

Vision-language models (VLMs) demonstrate emergent capabilities such as zero-shot open-vocabulary object detection~(\cref{sec:evals}), enabling them to localize and classify objects across unbounded semantic spaces. These capabilities prompt a fundamental question — \textit{How do VLMs adapt their mechanisms to perform zero-shot open-ended object detection?} In this section, we address this question by analyzing the layer dynamics of Grounding DINO~\cite{gdino} (VLM) and contrast them with DINO-DETR~\cite{detr} (vision-only object detector). We employ \textit{approximation probes} to analyze the information content and trajectories across the decoder layers. 

Our analysis reveals a phase transition~\cite{pt_dlst} in the decoder layers of both model types. In this critical reorganization phase, representations abruptly shift from generic features to task-specific compositional abstractions. This shared phenomenon underscores a striking similarity in the fundamental mechanisms utilized by these models and suggests that VLMs, analogous to vision-only object detectors, also employ hierarchical feature learning to perform object detection. Consequently, this shows that the emergent open-world capabilities of VLMs derive from their unique ability to integrate language-aligned representations into this hierarchy, enabling semantic flexibility. 

\subsection{Approximation Probe}

Linear representation hypothesis~\cite{weaklrh, linearprobes} states that representations in a deep neural network correspond to directions in the latent space and become increasingly linearly separable with depth, enabling simple linear models to measure the task-relevant information content at a particular layer. 

Formally, let \( \phi_\ell: \mathcal{X} \rightarrow \mathbb{R}^d \) denote the feature extractor up to layer \( \ell \), mapping an input image \( x \in \mathcal{X} \) to activations \( Z_\ell \in \mathbb{R}^d \). For a dataset \( \mathcal{D} = \{(x_i, y_i)\}_{i=1}^N \) with labels \( y_i = (y_{\text{class}}^i, y_{\text{bbox}}^i) \), we train two linear probes per layer as shown in \cref{fig:probes}:

\paragraph{Classification probe.} Predicts class labels \( y_{\text{class}} \in \mathcal{Y}_{\text{class}} \):
\begin{equation}
\min_{W_{\text{class}}, b_{\text{class}}} \sum_{i=1}^N \mathcal{L}_{\text{CE}}\left( W_{\text{class}} \cdot \phi_\ell(x_i) + b_{\text{class}}, y_{\text{class}}^i \right)
\end{equation}
where \( \mathcal{L}_{\text{CE}} \) is the cross-entropy loss, \( W_{\text{class}} \in \mathbb{R}^{|\mathcal{Y}_{\text{class}}| \times d} \) is the weight matrix, and \( b_{\text{class}} \in \mathbb{R}^{|\mathcal{Y}_{\text{class}}|} \) is the bias term.

\paragraph{Localization probe.} Predicts bounding box coordinates \( y_{\text{bbox}} \in \mathbb{R}^4 \):

\begin{equation}
\min_{W_{\text{bbox}}, b_{\text{bbox}}} \sum_{i=1}^N \mathcal{L}_{\text{sL1}}\left( W_{\text{bbox}} \cdot \phi_\ell(x_i) + b_{\text{bbox}}, y_{\text{bbox}}^i \right)
\end{equation}

where \( \mathcal{L}_{\text{sL1}} \) is the smooth L1 loss, \( W_{\text{bbox}} \in \mathbb{R}^{4 \times d} \) is the weight matrix, and \( b_{\text{bbox}} \in \mathbb{R}^4 \) is the bias term.

The probe accuracies \( \text{Acc}_{\text{class}}(\ell) \) and \( \text{Acc}_{\text{bbox}}(\ell) \), reported as average precision at IoU 50, measure how well \( Z_\ell \) encodes semantic and spatial information, respectively. To probe task-specific trajectories, we train the probes to approximate the raw predictions of the model. The accuracies achieved at each layer provides a quantitative lens into the information sufficiency of respective representations to perform the task under the assumption of linear representation hypothesis.

\subsection{Probe trajectories reveal phase transition}

Approximation probe trajectories across decoder layers reveal distinct phases of representation transformations in both VLMs and vision-only object detectors as shown in~\cref{fig:phase_transition}.

In the early layers ($\ell < \ell^*$), the accuracy of both the probes depicts lower performance than the final layer. As we move further into the intermediate layers, the accuracy drops until a transition layer ($\ell = \ell^*$), after which it surges to optimal task performance as we reach the late layers ($\ell > \ell^*$). This phenomenon, termed as \textit{phase transition}, has been previously observed in various scenarios such as training dynamics of deep neural networks~\cite{pt_train}, scale dynamics of diffusion models~\cite{pt_diffusion}, layer dynamics of vision transformers~\cite{pt_vit} etc., and has profound implications for generalization and hierarchical nature of the models. 

Both model types exhibit a similar phenomenon irrespective of the dataset or architecture. This indicates that VLMs utilize similar mechanisms as vision-only object detectors to perform the task of object detection. Further, while both tasks exhibit phase transitions, the prevalence for classification only arises for label sets with many classes. This could be attributed to the unnecessity of complex transformations for small label sets which contain sufficiently distinctive representations.

\subsection{Three-phase representation evolution mechanism}

As observed in~\cref{fig:phase_transition}, the models process the representations in three phases, which we propose as \textit{extraction phase}, \textit{reorganization phase} and \textit{refinement phase}. The universality of probe trajectories establishes the phenomena of phase transition as a fundamental property of hierarchical representation learning, a crucial component for generalization and composition in neural networks. This can be justified through the lens of information bottleneck (IB) principle~\cite{ibp} and random hierarchy model (RHM)~\cite{rhm} as further described in~\cref{appx:probes}.

Furthermore, as both model types demonstrate this phenomena, it is evident that the emergent capabilities demonstrated by VLMs are contributed primarily by the representational alignment of modalities, enforced by cross-modality fusion~\cite{gdino} or projection of visual tokens into the language space~\cite{florence, gemini}.

\section{Sparse Autoencoders for Concept Decomposition and Discovery}
\label{sec:concept}

\begin{figure*}[ht]
\begin{center}
\includegraphics[width=\textwidth]{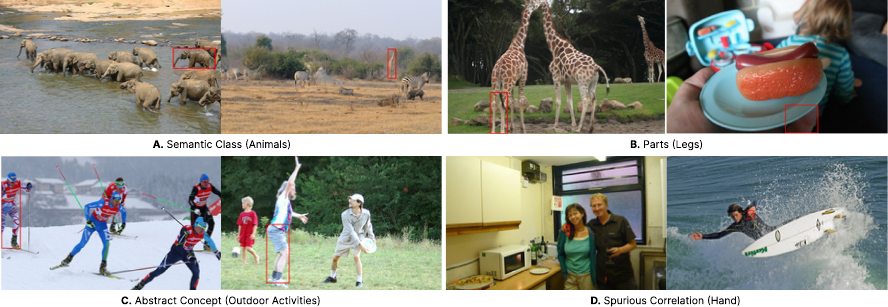}
\caption{Concepts discovered from TopK SAE~\cite{sae_topk} trained on features from Grounding DINO~\cite{gdino}. Note, that this shows only a subset of concepts and tokens per concept from a larger sample set.}
\vspace{-5mm}
\label{fig:concepts}
\end{center}
\end{figure*}

Sparse Autoencoders (SAEs) have recently emerged as a promising technique for overcoming the challenge of polysemanticity~\cite{sae_relu} and enables concept-based interpretability of models~\cite{sae_features}. As a valuable tool for mechanistic interpretability, SAEs represent an important component in the AI safety toolkit, enabling applications such as model steering~\cite{sae_steering}, mechanistic analysis~\cite{mechinterp, sae_refusal}, model debugging~\cite{probe_debug}, \etc~ Hence, we integrate several variants of SAEs, namely, ReLU~\cite{sae_relu}, TopK~\cite{sae_topk}, Batch TopK~\cite{sae_batchtopk} and Matryoshka~\cite{sae_matryoshka}, which can also be used for transcoding~\cite{sae_transcode} and crosscoding~\cite{sae_crosscode}.

In this section, we demonstrate that SAEs recover human-interpretable concepts for vision-language models on the task of object detection, focusing specifically on Grounding DINO~\cite{gdino}. Furthermore, to interpret the decomposed sparse units (basis vectors), we utilize \textit{dataset attribution} to discover human-interpretable concepts and present some representative examples. 

\subsection{Concept Decomposition}

To decompose the internal representations into a sparse set of interpretable codes, we train TopK SAE~\cite{sae_topk} on features extracted from the residual stream of the decoder layers of Grounding DINO~\cite{gdino}.

Formally, the architecture consists of two linear layers, an encoder $E$ and a decoder $D$. The encoder $E$ maps the normalized input feature vector $\mathbf{x}' \in \mathbb{R}^d$ to a higher-dimensional latent representation $\mathbf{z} = E(\mathbf{x}')$, where $\mathbf{z} \in \mathbb{R}^m$. The dimensionality of the latent space is defined as $m = d \times e$, where $e$ is the expansion factor. To enforce sparsity in the latent representation, a \textit{TopK} non-linearity is applied to the output of the encoder, resulting in a sparse code $\hat{\mathbf{z}} = \text{TopK}(\mathbf{z})$. The $\text{TopK}(\cdot)$ function sets all but the $k$ largest elements of $\mathbf{z}$ to zero for a given feature input. The decoder $D$ is a linear layer that maps this sparse code back to the normalized input space: $\hat{\mathbf{x}}' = D(\hat{\mathbf{z}})$, where the rows represent the sparse basis vectors representing individual concept.

The SAE is optimized with a total loss function $\mathcal{L}_{\text{total}}$, which is a weighted sum of a reconstruction loss ${L}_{\text{recon}}$ and an auxiliary loss $\mathcal{L}_{\text{aux}}$ as described in~\cite{sae_topk}.

\subsection{Concept Discovery}

To discover the concepts encoded by the sparse latent units (corresponding to the basis vectors of the decoder dictionary $B$), we employ the \textit{dataset attribution} method, which gathers the maximally activating feature from a dataset $D$, for each individual sparse unit $i \in B$. These top-activating inputs (patches or proposals) are then manually interpreted by a human. By examining the common visual characteristics across the top-activating examples for a given sparse unit, we can infer the concept that the unit represents and assign it an interpretable label. We use the validation set of COCO~\cite{coco} as dataset $D$ and visualize top 64 features per sparse unit $i$. For object detection and in this case with Grounding DINO~\cite{gdino}, each feature corresponds to an object as depicted by the bounding box in~\cref{fig:concepts}.

\paragraph{Interpretable sparse units.} Analyzing the sparse units from an SAE trained on features from the residual stream of decoder layer 4 of Grounding DINO~\cite{gdino} revealed several meaningful concepts relevant to the task of object detection as shown in~\cref{fig:concepts}. These include, \textit{semantic classes} such as units activated by specific visual patterns associated with \textit{animals} (A), \textit{abstractions} (C) such as features related to people playing in the ground or \textit{parts} (B) of objects, even though this supervision was not present in the training dataset of the object detector. 

\paragraph{Spurious correlations.} Importantly, this analysis also uncovered sparse units that appear to encode spurious correlations learned by the model, highlighting potential vulnerabilities and failure modes. A prominent example of such a spurious concept is a sparse unit that is strongly activated by proposals containing \textit{hands} (D). Dataset attribution for this \textit{hand} unit showed that it frequently co-activated with model predictions for objects commonly held in hands, such as knife (image 1 of D) or cell phone (image 2 of D), even when the objects themselves were absent or visually ambiguous in the image. This suggests that the model might, in some cases, rely on superficial contextual cues (like a hand) as a shortcut for prediction, rather than on robust object features.

\section{Conclusion}
\label{sec:conclusion}

In this work, we introduce \textsc{BlueGlass}, a framework designed to facilitate composite AI safety research by enabling the integration and composition of diverse safety tools, operating across diverse model aspects. Furthermore, we demonstrate the utility of this framework in supporting empirical and mechanistic safety workflows by performing three distinct safety-oriented analyses and methodologies applied to vision-language models (VLMs) for the task of object detection. The distributional evaluation highlights current performance trade-offs and potential failure modes of VLMs across diverse data distributions and operational conditions. Then, we propose approximation probes to analyze the layer dynamics of VLMs, which reveals a shared hierarchical feature learning principle between VLMs and vision-only detectors as evidenced by a shared phase transition phenomena. Lastly, we demonstrate the applicability of sparse autoencoders (SAEs) for concept decomposition and discovery on VLMs for the task of object detection. This identifies interpretable concepts learned by VLMs and uncovers spurious correlations which harm their reliability. Collectively, these findings empirically characterize VLMs' behavior and limitations which are crucial for ensuring their reliable and safety-aware deployments in real-world applications, as well as, for informing future research directions aimed at enhancing their robustness and performance.

\section{Impact Statement}

This work contributes tools for building, composing and scaling safety methodologies for AI systems and presents analyses for behavioral and mechanistic aspects of vision-language models for object detection through a safety perspective. We expect the contributions to accelerate empirical safety research and improve the transparency and reliability of models deployed in real-world settings.


\bibliography{main}
\bibliographystyle{icml2025}


\newpage
\appendix
\onecolumn

\section{Related works}
\label{sec:related_work}

\paragraph{Object detection.} A fundamental task in computer vision involves localization and classification of objects in a scene~\cite{odsurvey}. With several years of advancements, object detectors~\cite{rcnn, yolo, detr} have demonstrated human-level performance on multiple closed-set benchmarks. With the widespread usage of deep learning-based methods in various applications, object detection has shifted towards open-world settings~\cite{owd}, which aims to extend the closed-set setting to unbounded input spaces, a crucial capability for reliable real-world deployment. 

\paragraph{Vision-language models for object detection.}
Vision-language models (VLMs) have advanced object detection by integrating visual and textual information, enabling capabilities such as open-world detection, which is crucial for real-world scenarios. Early approaches~\cite{gdino,genu} used separate vision and language encoders, while recent models~\cite{florence,gemini} learn joint representations. VLMs employ diverse architectural approaches but can be encompassed into three broad categories, namely, contrastive with proposals~\cite{gdino}, generative with proposals~\cite{genu} and generative without explicit detection networks~\cite{florence, gemini}.

\paragraph{Mechanistic interpretability.} Mechanistic interpretability~\cite{mechinterp} aims to understand how neural networks process information structurally. Early work focused on feature visualization~\cite{featvis} and attribution methods~\cite{attribution}, while recent research explores circuit-level analysis~\cite{circuits} in models like GPT-2~\cite{gpt2} and vision transformers~\cite{vit}. The field has developed several tools, such as linear probes~\cite{linearprobes}, activation patching~\cite{actpatch}, sparse autoencoders~\cite{sae_relu}, \etc~to investigate the internal mechanisms of a model. This growing field enhances transparency, aiding in debugging, robustness, and alignment of AI systems.
\section{Design Principles}
\label{appx:principles}

The following principles guide the development of the system to ensure it remains adaptable, efficient, and suitable for a wide range of safety research workflows with ease.

\paragraph{Generality.} In practice, system components vary across environments in architecture, implementation style and complexity. To ensure broad applicability, we should avoid dependence on any framework specifics or design patterns. Instead, lightweight and extensible abstractions expose functionality through uniform interfaces, enabling flexibility and consistency across components.

\paragraph{Composability.} Reliable oversight often requires combining multiple tools such as probes, interpreters, evaluators, and patchers into coordinated workflows. To facilitate this, all components should be designed as modular, interoperable units that operate over shared data formats and can be flexibly composed without any significant instrumentation effort.

\paragraph{Resourcefulness.} Modern AI systems are characterized by the immense scale of models and datasets. Consequently, safety techniques that operate on model internals and on comprehensive datasets should efficiently utilize resources through caching, selective instrumentation and lazy evaluation, while integrating cleanly with workflows such as distributed pipelines, batched operations and asynchronous execution.   

\paragraph{Usability.} The primary goal of the framework is to enable ease-of-use and simplified integration of AI safety tools and workflows. This requires abstracting away the underlying complexities of interacting with models, data formats, safety methods, and providing a high-level interface that promotes experimentation and reliable deployment of composite AI safety workflows.    
\section{Categorization of Vision-Language Model Architectures}
\label{appx:vlm_arch}

\begin{figure}[ht]
\begin{center}
\centerline{\includegraphics[width=0.9\textwidth]{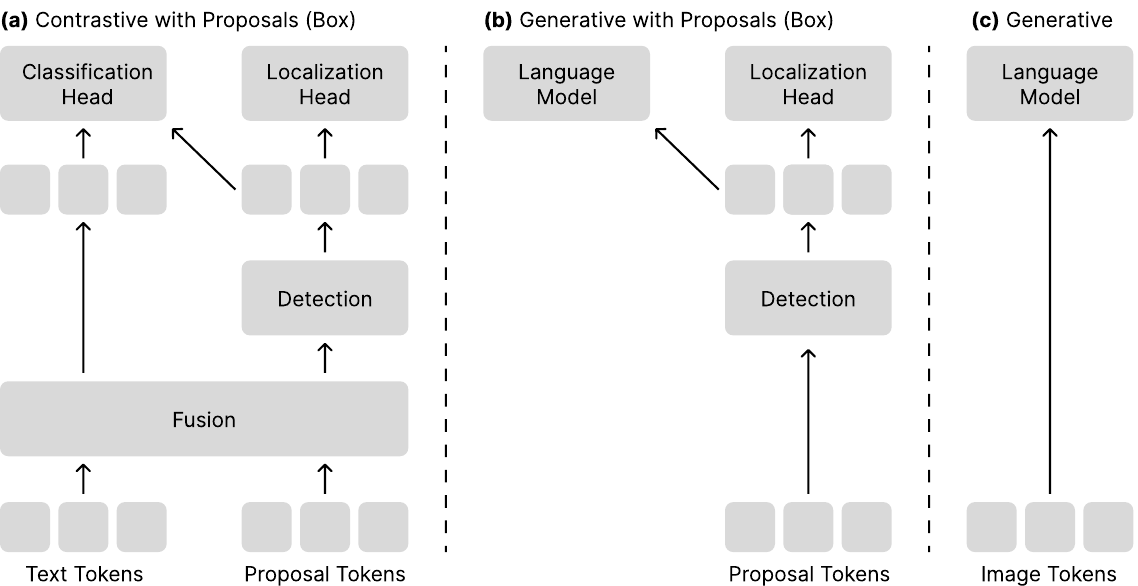}}
\caption{Categorization of Vision-language model based on architecture and learning objective.}
\label{fig:vlm_arch}
\end{center}
\end{figure}

Vision-Language Models (VLMs) have emerged with diverse architectural paradigms that dictate how visual and linguistic information are integrated and processed. These architectural choices fundamentally influence the VLM's capabilities and its suitability for various multimodal tasks. For the purpose of understanding their application in object detection and other safety-critical analyses, VLMs can be broadly categorized into three core architectural classes based on their primary learning objective and output mechanism as shown in~\cref{fig:vlm_arch}.

\paragraph{Discriminative vision-only detectors.} Traditionally, object detectors are designed to directly map the image inputs to specific localization and classification predictions. Their primary objective is to learn a decision boundary for each of these tasks. These models typically operate in a supervised learning setting, where they are trained with labeled data to predict discrete classes and bounding boxes. In our evaluation analysis, these correspond to the baseline~\cite{yolo} and the fine-tuned upperbound~\cite{detr}.

\paragraph{Contrastive VLMs.} VLMs in this architectural class learn to align representations from different modalities (e.g., images and text) into a shared, semantically rich latent space. Their learning objective is to maximize the similarity between positive pairs (e.g., an image and its matching text description) and minimize the similarity between negative pairs. For object detection, these models typically utilize specialized networks for localization predictions and perform classification by learning to contrast between a predefined set of classes and the textual predictions. Architecturally, contrastive VLMs commonly employ dual-encoder (two-stream) designs, where separate encoders process each modality independently, followed by a projection mechanism to align their embeddings. In our analysis, this corresponds to Grounding DINO~\cite{gdino}.

\paragraph{Generative VLMs.} Generative VLMs are focused on producing new content in one modality based on input from another, or generating multimodal outputs from a given prompt. Their objective is to learn the underlying data distribution to create novel, coherent, and contextually relevant outputs. This category encompasses tasks such as image captioning (generating text from images), text-to-image generation, or multimodal dialogue systems. Architecturally, generative VLMs often utilize encoder-decoder structures, where this category can be further divided into (A) models that utilize proposal networks for encoder; and (B) models that utilize a simpler patch-based visual encoder. The encoded features (tokens) are then fed into the decoder, to generate the required outputs. In our analysis, we evaluate two models, Gemini 2.0 Flash~\cite{gemini} (closed source) and Florence 2~\cite{florence} (open source), that belong to category B and one model from category A, namely GenerateU~\cite{genu}.

The presented models are the best performing in their individual categories and are representative of the behavior of other models in the same category. Hence, we only consider these models for our analyses.

\section{Evaluation Pipeline}
\label{appx:eval_ablation}

\begin{figure}[ht]
\begin{center}
\centerline{\includegraphics[width=\textwidth]{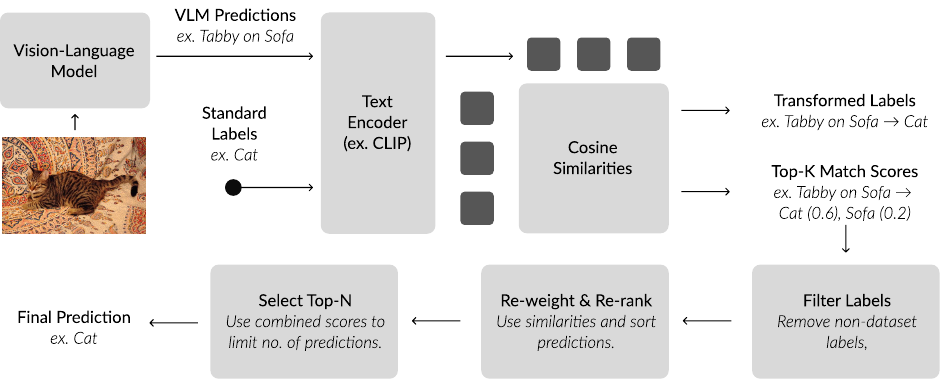}}
\caption{Evaluation pipeline.}
\label{fig:eval_pipeline}
\end{center}
\end{figure}

The open-ended nature of VLM output space poses challenges for standard evaluation schemes where a fixed set of class labels are expected. To address this, we propose an evaluation pipeline, as shown in~\cref{fig:eval_pipeline}, that maps VLM outputs to a predefined label space using a semantic (text) similarity as also proposed in some prior works~\cite{embedsim, genu}. 

\begin{table}[ht]
\centering
\begin{tabular}{lcccccccccc}
\toprule
Group  & Encoder & Max Pred & Min Conf & Objn. & TopK & Negatives & Parts & COCO mini (AP)  \\
\midrule
Baseline & CLIP & 900 & 0 & $\times$ & $\times$ & $\times$ & $\times$ & 27.6 \\
\midrule
Use Box & CLIP & 900 & 0 & $\checkmark$ & $\times$ & $\times$ & $\times$ & 42.8 \\
\midrule
Use TopK & CLIP & 900 & 0 & $\checkmark$ & $\checkmark$ & $\times$ & $\times$ & 42.8 \\
\midrule
Use Neg & CLIP & 900 & 0 & $\checkmark$ & $\checkmark$ & $\checkmark$ & $\times$ & 44.6 \\
\midrule
Use Parts & CLIP & 900 & 0 & $\checkmark$ & $\checkmark$ & $\checkmark$ & $\checkmark$ & 44.5 \\
\midrule
Max Pred & CLIP & 10 & 0 & $\checkmark$ & $\checkmark$ & $\checkmark$ & $\checkmark$ & 40.9 \\
 & CLIP & 30 & 0 & $\checkmark$ & $\checkmark$ & $\checkmark$ & $\checkmark$ & 43.1  \\
 & CLIP & 100 & 0 & $\checkmark$ & $\checkmark$ & $\checkmark$ & $\checkmark$ & 44.3  \\
 & CLIP & 300 & 0 & $\checkmark$ & $\checkmark$ & $\checkmark$ & $\checkmark$ & 44.4 \\
 & CLIP & 900 & 0 & $\checkmark$ & $\checkmark$ & $\checkmark$ & $\checkmark$ & 44.5  \\
\midrule
 Min Conf & CLIP & 900 & 0 & $\checkmark$ & $\checkmark$ & $\checkmark$ & $\checkmark$ & 44.5 \\
 & CLIP & 900 & 0.5 & $\checkmark$ & $\checkmark$ & $\checkmark$ & $\checkmark$ & 28.4  \\
 & CLIP & 900 & 0.1 & $\checkmark$ & $\checkmark$ & $\checkmark$ & $\checkmark$ & 43.8 \\
 & CLIP & 900 & 0.01 & $\checkmark$ & $\checkmark$ & $\checkmark$ & $\checkmark$ & 44.5 \\
 & CLIP & 900 & 0.001 & $\checkmark$ & $\checkmark$ & $\checkmark$ & $\checkmark$ & 44.5 \\
\midrule
Encoder & BERT & 900 & 0 & $\checkmark$ & $\checkmark$ & $\checkmark$ & $\checkmark$ & 27.7 \\
 & B1ADE & 900 & 0 & $\checkmark$ & $\checkmark$ & $\checkmark$ & $\checkmark$ & 36.7 \\
 & CLIP & 900 & 0 & $\checkmark$ & $\checkmark$ & $\checkmark$ & $\checkmark$ & 44.5  \\
 & NVEMBED & 900 & 0 & $\checkmark$ & $\checkmark$ & $\checkmark$ & $\checkmark$ & 43.7 \\
\bottomrule
\end{tabular}
\caption{Ablations of control parameters and components for evaluation pipeline on predictions from GenerateU~\cite{genu} on COCO~\cite{coco} mini dataset. Explanation for each group and parameters are detailed in~\cref{sec:mech}. Best configuration corresponds to higher average precision (AP) scores. }
\label{tab:eval_ablations}
\end{table}

Let $f_{\text{encoder}}: \mathbb{T} \rightarrow \mathbb{R}^d$ denotes a text encoder, $\mathcal{C} = \{c_1, c_2, \dots, c_N\}$ represents the fixed set of $N$ dataset-specific class labels, and \(\mathcal{O} = \{o_1, o_2, \dots, o_M\}\) represent the \(M\) open-ended textual outputs from the VLM. We encode $\mathcal{C}$ and $\mathcal{O}$ using the text encoder $f_{\text{encoder}}(\cdot)$

\begin{equation}
    \mathbf{e}_{c_i} = f_{\text{encoder}}(c_i), \quad \mathbf{e}_{o_j} = f_{\text{encoder}}(o_j),
\end{equation}

where $\mathbf{e}_{c_i}, \mathbf{e}_{o_j} \in \mathbb{R}^d$ are the text embeddings, and $d$ is the embedding dimension. Then, for each VLM output $o_j$, we compute pairwise similarities with all $c_i \in \mathcal{C}$

\begin{equation}
    \text{sim}(o_j, c_i) = \mathbf{e}_{o_j}^\top \mathbf{e}_{c_i}.
\end{equation}
    
and assign the transformed label $\hat{y}_j$ to $o_j$ such that
    \begin{equation}
        \hat{y}_j = \underset{c_i \in \mathcal{C}}{\arg\max} \, \text{sim}(o_j, c_i).
    \end{equation}

\subsection{Mechanisms for Robustness}
\label{sec:mech}

Vision-Language Models (VLMs), particularly in open-vocabulary settings, can generate a vast range of predictions that may not align with the specific ground truth label set used for evaluation. This leads to two primary challenges: outputs that are ungrounded or irrelevant to the desired set of detectable objects, and hierarchical semantic ambiguities where parts of an object are mapped as the whole. Our methodology incorporates specific mechanisms to address these issues.

\paragraph{Filtering irrelevant predictions with negative classes.} VLM outputs are often highly granular and can include predictions for concepts that do not correspond to any desired label in the evaluation set, or are simply ungrounded. To mitigate these irrelevant outputs, we introduce \textit{negative classes}. This involves augmenting the candidate label set with additional prompts, such as "an object" or "a thing," representing more general or undesirable categories. VLM predictions tend to align more strongly with these negative prompts when the ground truth label is absent and are effectively filtered out, ensuring that only more relevant and grounded predictions are considered for evaluation. The ablations presented in~\cref{tab:eval_ablations} demonstrate that the inclusion of negative classes significantly improves precision by reducing false positives from ungrounded or out-of-scope predictions.

\paragraph{Resolving part-to-whole ambiguities with part prompts.} Another challenge in evaluating VLM predictions with the proposed pipeline is the misclassification of object parts as complete objects (e.g., a "wheel" being incorrectly identified as a "car"). To handle this part-to-whole ambiguity, we incorporate specific prompts for object parts. For each relevant object class, we augment the text query space with prompts structured as "parts of {class name}" (e.g., "parts of a car"). By providing these explicit part-level prompts, the text encoder is guided to differentiate between a whole object and its constituent components, leading to more accurate and contextually appropriate detections. Our ablations indicate the positive impact of this strategy on detection performance, as shown in~\cref{tab:eval_ablations}.

\subsection{Control Parameters and Components}

Beyond the strategies for handling open-vocabulary ambiguities, the overall performance of VLM-based detection systems is highly influenced by various control parameters and the choice of encoder. We conduct extensive ablation studies to analyze the impact of critical hyperparameters such as prediction confidence thresholds (\textit{Min Conf}), maximum number of predictions (\textit{Max Pred}), inclusion of objectness scores in final scores (\textit{Objn.}) and the selection of different text encoder models (CLIP~\cite{clip}, NVEmbed~\cite{nvembed}, B1ade~\cite{blade} and BERT~\cite{bert}). The detailed results of these ablations, demonstrating their individual and combined effects on performance, are presented in~\cref{tab:eval_ablations}.

\section{Phase Transition and Hierarchical Feature Learning}
\label{appx:probes}

The observed phase transition phenomenon in the decoder layers of VLM and vision-only object detectors is a consequence hierarchical feature learning within neural networks. This phenomenon, justified through the lens of the Information Bottleneck (IB) principle~\cite{ibp} and Random Hierarchy Models (RHM)~\cite{rhm}, highlights how models iteratively refine and compose representations, shedding irrelevant information to form increasingly abstract features.

\paragraph{Information bottleneck principle.}
The information bottleneck (IB) principle~\cite{ibp} formalizes the trade-off between compressing input data $X$ and preserving task-relevant information $Y$. Let $Z_\ell$ denote the activations at layer $\ell$. The IB objective is can be denoted as

\begin{equation}
L_{\text{IB}} = I(Z_\ell; Y) - \beta I(Z_\ell; X),
\end{equation}

where $I(Z_\ell; X)$ and $I(Z_\ell; Y)$ represent the mutual information between features, inputs and outputs, and $\beta > 0$ controls the compression-relevance trade-off. Early layers maximize $I(Z_\ell; X)$ to extract low-level or generic features, while deeper layers minimize $I(Z_\ell; X)$ to discard irrelevant details and maximize $I(Z_\ell; Y)$.

\begin{proposition}[Phase transition in mutual information]
\label{prop:mi} 
In a hierarchical network, there exists a critical layer $\ell^*$ where $I(Z_\ell; X)$ undergoes a sharp decrease (compression), and $I(Z_\ell; Y)$ exhibits a non-monotonic trajectory (reorganization dip). This corresponds to a phase transition from generic to task-specific representations.
\end{proposition}

\paragraph{Random hierarchy models (RHM).} The random hierarchy model assumes that the data is structured as a tree, where coarse features (e.g., class labels) depend on fine features (e.g., edges, textures) and the model iteratively resolves ambiguities at each level while composing the representation into more abstract forms. 

\begin{proposition}[Phase transition in hierarchical models]
\label{prop:hm}
Let $H$ denote the feature hierarchy, $H = \{\phi_1, \phi_2, \ldots, \phi_L\}$, where $\phi_\ell$ represents features at layer $\ell$. Compositionality requires that $\phi_{\ell+1} = g(\phi_\ell)$, where $g$ is a non-linear function resolving ambiguities (e.g., pooling edges into shapes). Linear transformations cannot resolve compositional ambiguities. Let $\phi_\ell \in \mathbb{R}^d$ and $\phi_{\ell+1} = W \phi_\ell$. If $W$ is linear, then $\phi_{\ell+1}$ inherits the ambiguities of $\phi_\ell$. Non-linear $g$ (e.g., ReLU) is necessary to discard irrelevant activations and compose features.
\end{proposition}

\begin{theorem}[Non-linearity as Phase transition]
If $f_\ell$ is linear for all $\ell$, then $I(Z_\ell; X) = I(X; Y)$ for all $\ell$, and no compression or composition occurs. Hence phase transition is a fundamental property of hierarchical feature learning.
\end{theorem}

The non-linearity in the probe trajectories and hence the phase transition (reorganization dip) corresponds to \cref{prop:mi} and \cref{prop:hm}. Non-linearity enables the network to discard irrelevant information and resolve ambiguities to compose hierarchical representations.


\end{document}